\documentclass[conference]{IEEEtran}
\IEEEoverridecommandlockouts

\usepackage{times}
\usepackage{float}

\usepackage{cite}

\usepackage{multicol}
\usepackage{amsmath,amssymb,amsfonts}
\usepackage{algorithmic}
\usepackage{graphicx}
\usepackage{textcomp}
\usepackage{xcolor}

\usepackage{geometry}

\usepackage{tabularx,booktabs}
\newcolumntype{C}{>{\centering\arraybackslash}X}
\begin{document}

\title{From Words to Poses: Enhancing Novel Object Pose Estimation with Vision Language Models\\}

\author{
    \IEEEauthorblockN{Tessa Pulli\IEEEauthorrefmark{1}, Stefan Thalhammer\IEEEauthorrefmark{2}, Simon Schwaiger\IEEEauthorrefmark{2}, Markus Vincze\IEEEauthorrefmark{1}}
    \IEEEauthorblockA{\IEEEauthorrefmark{1}Vision for Robotics Laboratory, Automation and Control Institute, TU Wien, Austria\\
    \{pulli, vincze\}@acin.tuwien.ac.at}
    \IEEEauthorblockA{\IEEEauthorrefmark{2}Industrial Engineering Department, UAS Technikum Vienna, Austria\\
   \{stefan.thalhammer, schwaige\}@technikum-wien.at}
}

\newgeometry{top=72pt, left=54pt, right=54pt, bottom=54pt}
\maketitle

\begin{abstract}
Robots are increasingly envisioned to interact in real-world scenarios, where they must continuously adapt to new situations. 
To detect and grasp novel objects, zero-shot pose estimators determine poses without prior knowledge.
Recently, vision language models (VLMs) have shown considerable advances in robotics applications by establishing an understanding between language input and image input. 
In our work, we take advantage of VLMs zero-shot capabilities and translate this ability to 6D object pose estimation.
We propose a novel framework for promptable zero-shot 6D object pose estimation using language embeddings.
The idea is to derive a coarse location of an object based on the relevancy map of a language-embedded NeRF reconstruction and to compute the pose estimate with a point cloud registration method.
Additionally, we provide an analysis of LERF's suitability for open-set object pose estimation.
We examine hyperparameters, such as activation thresholds for relevancy maps and investigate the zero-shot capabilities on an instance- and category-level. 
Furthermore, we plan to conduct robotic grasping experiments in a real-world setting.
\end{abstract}

\section{Introduction}

\begin{figure*}[!t]
    \centering
    \def\svgwidth{\textwidth}
    \includegraphics[width=1\textwidth]{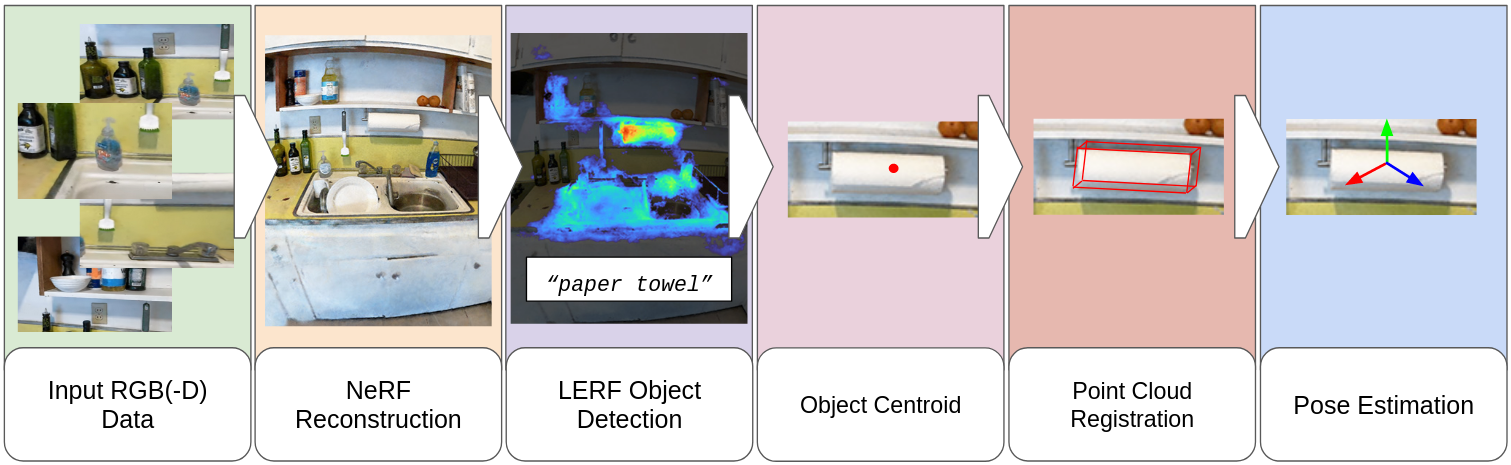}
    \caption{From a set of RGB(-D) images, a NeRF scene is reconstructed. Using LERF, the target object is detected via text prompting. The object centroid is then computed through three-dimensional semantic segmentation. Finally, the pose estimate is determined using a point cloud registration method, e.g. Teaser++~\cite{yang2020teaser}.}
    \label{fig:overview_pipeline}
  \end{figure*}
\vspace{-2mm}

6D object pose estimation of unseen objects is a core task in robotics.
Classical methods estimate the pose of objects using trained networks either for object instances \cite{wang2021gdr, su2022zebrapose, park2019pix2pose} or zero-shot methods with the idea of capturing classes of objects and adapting to novel objects~\cite{nguyen2024gigapose, labbe2022megapose, lin2024sam6dsegmentmodelmeets}. 
With the advent of vision language models (VLM)~\cite{radford2021clip, jia2021align}, novel methods to detect objects and align visual content with natural language show noteworthy results for realistic scenes.  
Traditional pose estimation methods output the required pose to manipulate objects in a real-world scene without any context.
VLMs show impressive results in terms of object recognition \cite{jin2024llms, zang2023contextual}, scene understanding, and even reasoning \cite{fu2024scene, ha2022semantic}.
We propose to take advantage of the recent advances in VLMs and utilize their zero-shot capability in 6D object pose estimation methods by introducing a novel promptable zero-shot 6D object pose estimation pipeline.
We explore VLMs for open-vocabulary object pose estimation, leveraging their zero-shot scene understanding capabilities~\cite{radford2021clip}. 
Using NeRF and the language embedding LERF~\cite{kerr2023lerf}, we query objects in an open-vocabulary manner.
The LERF-generated relevancy map provides the object's location from which the centroid in 3D space can be derived. 
The 6D pose is estimated using a point cloud registration method like TEASER++~\cite{yang2020teaser}. Finally, grasp points and affordances are derived for real-world manipulation.
In summary, the paper has the following key contributions:
\begin{itemize}
    \item We introduce a language-embedded zero-shot object pose estimation framework.
    \item We analyze the zero-shot capabilities of LERF to identify key requirements to enhance their applicability in pose estimation.
\end{itemize}

\section{Related works}
Several works incorporate VLMs in robotics-related scenarios with considerable results~\cite{song2024socially, shah2023lm, deng2024openobj, qiu2024learning}.
~\cite{song2024socially, shah2023lm} prove the potential of VLMs in robotics scenarios by also considering 3D data for promptable navigation.
Other works~\cite{qiu2024learning, deng2024openobj} introduce and perform robotic grasping experiments with VLM-based pipelines but avoid the estimation of object poses. 
Instead, GEFF~\cite{qiu2024learning} simplifies 6D object pose estimation by grasping the object's centroid, while~\cite{deng2024openobj} relies on an general object grasping methods~\cite{fang2020graspnet}, and~\cite{gao2023physically} replays previously collected pick-and-place primitives.
The mentioned approaches avoid 6D object pose estimation by using alternative strategies for object manipulation.
This simplification may lead to major limitations when it comes to manipulating objects with more complex shapes.
In this work, we propose a VLM-based method for estimating the 6D pose of novel objects to enable open-set manipulation in unknown settings.

\section{Method}

Figure~\ref{fig:overview_pipeline} illustrates our proposed pipeline. 
We reconstruct a scene based on a set of RGB(-D) images input by using NeRFstudio~\cite{nerfstudio}. 
We assume the availability of multiple images of a scene without camera poses and retrieve these using multiview stereo, e.g. COLMAP~\cite{schoenberger2016mvs}.
Having individual images and corresponding poses allows for a joint geometric reconstruction of the scene.
Within this framework, we use a CLIP-based language embedding~\cite{kerr2023lerf, radford2021clip} to query objects in an open-vocabulary manner. 

Through the relevancy map generated by the LERF response (see Fig. \ref{fig:lerf_output}), we can obtain a coarse 3D location of the target object.
Qui et al. \cite{qiu2024learning} estimate poses in their work by aggregating the semantic point cloud and calculating the centroid of the object. 
We want to take advantage of this approach and obtain a coarse location of the object based on its centroid.
Afterwards, the 6D pose of an object is estimated using a point cloud registration method, like Teaser++~\cite{yang2020teaser}.

\subsection{Promptable Object Localization}
\begin{figure}
    \centering
    \includegraphics[width=\columnwidth]{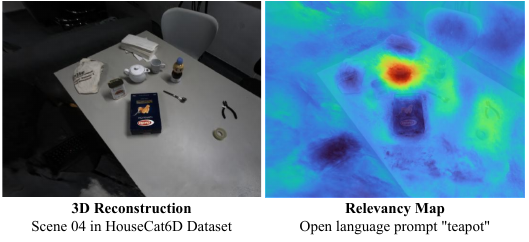}
    \caption{\textbf{VLM-based scene reconstruction.} 3D reconstruction of scene 04 of the HouseCat6D dataset \protect\cite{jung2024housecat6d} with overlaid relevancy map generated with LERF \protect\cite{kerr2023lerf} for open language prompt \textit{teapot}. Red shading indicates high relevancy between scene and prompt.}
    \label{fig:lerf_output}
\end{figure}

To validate the potential of LERF~\cite{kerr2023lerf} for 6D object pose estimation, we perform an analysis to reveal strengths and limitations of the method.
For our validation, we consider the performance of instance-level and category-level prompts and investigate with which language input the most promising results can be achieved. 
Furthermore, we plan to investigate hyperparameters, such as activation thresholds for the relevancy maps to understand how an optimal object centroid can be obtained. 
As we hypothesize that our method provides versatile applicability for household robotics, we test our approach on the dataset HouseCat6D~\cite{jung2024housecat6d}, which provides $41$ scenes with $194$ object instances. 
We reconstruct the scenes of the dataset with NeRFstudio~\cite{nerfstudio} and analyze the capabilities with the intention of utilizing the approach for 6D object pose estimation.
Figure \ref{fig:lerf_output} shows an exemplary HouseCat6D scene and relevancy map generated with a LERF language prompt.

\subsection{Promptable Object Pose Estimation}
To localize the target object, we use an approach similar to \cite{qiu2024learning}. 
Firstly, we filter point clouds for relevant vertex clusters with the CLIP activation.
Subsequently, the relevant pixels are separated into individual object instances with a clustering approach like DBScan \cite{ester1996dbscan}.
Based on the clustered point cloud, the object's centroid and, therefore, its coarse location are determined. 
Ultimately, the 6D object poses are estimated using RGB-D registration assuming the availability of the object's 3D mesh.
A modified version of TEASER++ \cite{yang2020teaser} registers an object prior with the observed partial point cloud, also accounting for the object texture.
This consideration is crucial as it allows to disambiguate geometric symmetries with texture cues enabling robust pose estimation.
Based on our proposed VLM-based 6D object pose estimation pipeline, we plan to conduct grasping experiments in a real-world household setting using HOPE and YCB-Video objects. 

\section{Discussion and Future work}
Future research will explore the applicability of zero-shot VLMs in settings beyond household environments, with a particular focus on industrial contexts. 
We believe our method holds significant potential for these settings, despite the considerable differences between industrial scenes and the datasets on which CLIP is pre-trained~\cite{radford2021clip}. 
One limitation of this study is the assumption that object priors are available. 
To address this, future work will aim to overcome this constraint by investigating derived affordances, enabling pose estimation and grasping without the need for pre-existing object models.

{\small
\bibliographystyle{IEEEtran}
\bibliography{IEEEabrv,root}
}

\end{document}